\documentclass[10pt,twocolumn,letterpaper]{article}

\usepackage[pagenumbers]{cvpr} 

\usepackage{graphicx}
\usepackage{amsmath}
\usepackage{amssymb}
\usepackage{booktabs}

\newcommand{\tickyes}{\checkmark}
\newcommand{\tickno}{\hspace{1pt}\ding{55}}
\usepackage{xcolor}
\usepackage[export]{adjustbox}
\usepackage{cellspace, tabularx}
 
\usepackage{caption}
\usepackage{subcaption}
\usepackage{stackengine}
\usepackage{cuted}

\usepackage{siunitx}
\usepackage{soul} 

\usepackage{floatrow}
\usepackage{multirow}
\newfloatcommand{capbtabbox}{table}[][\FBwidth]
\usepackage{blindtext}
\usepackage{subcaption} 
\usepackage{color, colortbl}
\definecolor{LightCyan}{rgb}{0.88,1,1}
\definecolor{Gray}{gray}{0.9}

\newcommand{\xb}{\bm{x}}

\newcommand{\LL}{\mathcal{L}}
\newcommand{\LLI}{\mathcal{L}^{\mathbb{I}}}
\newcommand{\LLE}{\mathcal{L}^{\mathbb{E}}}
\newcommand{\CE}{\LL_{CE}}
\newcommand{\KL}{\LL_{KL}}

\newcommand {\sbr}[1]{\left[#1\right]}

\newcommand{\argmin}{\operatornamewithlimits{arg~min}}
\newcommand{\argmax}{\operatornamewithlimits{arg~max}}

\usepackage{bm}

\usepackage{pifont}
\usepackage{enumitem}
\usepackage{amsmath}

\newlist{todolist}{itemize}{2}
\setlist[todolist]{label=$\square$}
\usepackage{pifont}
\usepackage[pagebackref,breaklinks,colorlinks]{hyperref}

\usepackage[capitalize]{cleveref}
\crefname{section}{Sec.}{Secs.}
\Crefname{section}{Section}{Sections}
\Crefname{table}{Table}{Tables}
\crefname{table}{Tab.}{Tabs.}

\usepackage[english]{babel}
\usepackage[utf8]{inputenc}
\usepackage{amsmath}
\usepackage{amsfonts}
\usepackage{graphicx}
\usepackage{algorithm}
\usepackage{algpseudocode}
\usepackage{soul}
\usepackage[accsupp]{axessibility}
\def\cvprPaperID{4686} 
\def\confName{CVPR}
\def\confYear{2022}

\begin{document}

\title{Not All Relations are Equal:\\ Mining Informative Labels for Scene Graph Generation}


\author{\parbox{16cm}{\centering
    
    {\large Arushi Goel$^1$, Basura Fernando$^{2}$, Frank Keller$^{1}$, and Hakan Bilen$^{1}$}\\
    {\normalsize
    $^1$School of Informatics, University of Edinburgh, UK \\
    $^2$CFAR, IHPC, A*STAR, Singapore.}}
}

\maketitle

\begin{abstract}

Scene graph generation (SGG) aims to capture a wide variety of interactions between pairs of objects, which is essential for full scene understanding. 
Existing SGG methods trained on the entire set of relations fail to acquire complex reasoning about visual and textual correlations due to various biases in training data. Learning on trivial relations that indicate generic spatial configuration like `on' instead of informative relations such as `parked on' does not enforce this complex reasoning, harming generalization. 
To address this problem, we propose a novel framework for SGG training that exploits relation labels based on their informativeness.
Our model-agnostic training procedure imputes missing informative relations for less informative samples in the training data and trains a SGG model on the imputed labels along with existing annotations.
We show that this approach can successfully be used in conjunction with state-of-the-art SGG methods and improves their performance significantly in multiple metrics on the standard Visual Genome benchmark. Furthermore, we obtain considerable improvements for unseen triplets in a more challenging zero-shot setting.
\end{abstract}

\section{Introduction}
\label{sec.intro}

In this paper, we look at a structured vision-language problem, scene graph generation \cite{krishna2017visual, xu2017scene}, which aims to capture a wide variety of interactions between pairs of objects in images. 
SGG can be seen as a step towards comprehensive scene understanding and benefits several high-level visual tasks such as object detection/segmentation \cite{ren2015faster, girshick2015fast, he2017mask}, image captioning \cite{lin2014microsoft, anderson2018bottom, huang2019attention, goel2020injecting}, image/video retrieval \cite{faghri2017vse++}, and visual question answering \cite{antol2015vqa, lu2016hierarchical}. 
In the literature, SGG is typically formulated as predicting a triplet of a localized subject-object pair connected by a relation (\eg \emph{person \textbf{wear} shirt}). 
Broadly, recent advances in SGG have been obtained by extracting local and global visual features in convolutional neural networks \cite{simonyan2014very, zhang2019graphical, hung2019union} or graph neural networks \cite{yang2018graph, xu2017scene, lin2020gps} combined with language embeddings \cite{mikolov2013distributed, pennington2014glove} or statistical priors \cite{zellers2018neural} for predicting relations between objects. 


\begin{figure}
\begin{center}
\includegraphics[width=0.95\linewidth]{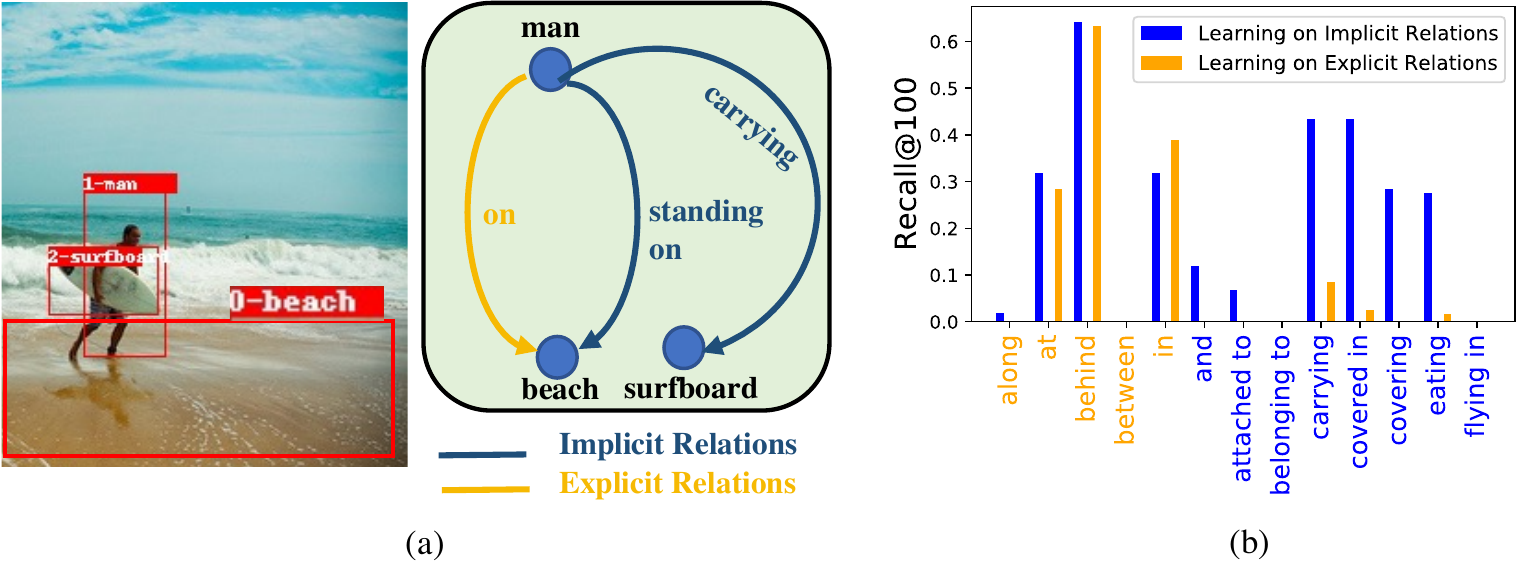}
\end{center}
\caption{(a) An example of a scene graph with \textit{implicit} and \textit{explicit} relations. (b) Per-class Recall@100 for an SGG model trained either on only explicit (orange) or implicit (blue) relations.}
\label{fig:intro-fig}
\end{figure}

Despite the remarkable progress in this task, various factors (long-tail data distribution, language or reporting bias~\cite{misra2016seeing}) in the established SGG benchmarks (\eg Visual Genome~\cite{krishna2017visual}) have been shown to drive existing methods towards biased and inaccurate relation predictions \cite{tang2018learning, tang2020unbiased}. 
One major cause is that each subject-object pair is annotated with \emph{only one positive} relation, which typically depends on the annotator's preference and is \emph{subjective}, while other plausible relations are treated as negative.\footnote{Note that both training and test sets of the standard benchmarks are subject to the similar biases.}
For instance, a subject-object pair \textit{man--beach} in \cref{fig:intro-fig}a is only annotated with one relation as \emph{man \textbf{on} beach}, even though other plausible relations are available such as \textit{standing on}.
Hence, the models \cite{zellers2018neural, hung2019union, tang2018learning, li2017scene} trained on this data become biased towards more frequently occurring labels, as reported in \cite{tang2020unbiased}.
To alleviate the biased training, Tang~\etal~\cite{tang2020unbiased} employ counterfactual causality to force the SGG model to base its predictions only on visual evidence rather than the data bias.
Wang~\etal~\cite{wang2020tackling} propose a semi-supervised technique that jointly imputes the missing labels of subject--object pairs with no annotated relations to obtain more balanced triplet distributions.
Suhail~\etal~\cite{suhail2021energy} propose an energy based method that can learn to model the joint probability of triplets from few samples and thereby avoids generating biased graphs with inconsistent structure.

While the recent methods~\cite{tang2020unbiased,wang2020tackling,suhail2021energy,chiou2021recovering} successfully tackle the bias towards more frequently occurring labels, this paper studies another type of bias related to \emph{label informativeness}. It also manifests itself in missing annotations and has not been addressed in SGG before.
In particular, we hypothesize that certain relation labels (implicit labels) are more informative than others (explicit labels) and training on implicit labels improves a model's ability to reason over complex correlations in visual and textual data.

Our key intuition comes from prior computational and cognitive models~\cite{logan1996computational} and recent work~\cite{collell2018acquiring} that categorize relations into \textit{explicit (or spatial)} and \textit{implicit}, based on whether the relation defines the relative spatial configuration between the two objects implicitly or explicitly (\eg \textit{man \textbf{standing on} beach vs.\ man \textbf{on} beach} in \cref{fig:intro-fig}a).
Explicit relations are often \emph{easy to learn}, \eg from the spatial coordinates of subject--object pairs, thanks to their highly deterministic spatial arrangements, while implicit ones are often challenging due to the relative spatial variation and require deliberate reasoning. 
To test our hypothesis, we conduct experiments where we train a SGG model either only on explicit, or only implicit relations, and evaluate them on a test set including both types (\ie zero-shot implicit or explicit relation classification), see \cref{fig:intro-fig}b.
Surprisingly, training only on implicit relations obtains good performance not only over implicit ones but also unseen explicit ones (only 2\% lower in average training on explicit relations and 4\% lower when trained on all labels), while training only on explicit relations performs poorly on implicit relations (where the performance drops to 0.1\% from 24.3\%).\footnote{We provide the full analysis in the results section and supplementary.}
In other words, training on implicit labels enables the model to better generalize to unseen explicit labels.
However, due to partially annotated training data, many subject-object pairs are only labelled by explicit relations and their implicit relations are missing and obscured by the explicit ones.

Motivated by our analysis, we design a novel model-agnostic training procedure for SGG that jointly extracts more information from partially labeled data by mining the missing implicit labels, trains a SGG model on them and boosts its performance.
In particular, our method involves a two stage training pipeline.
The first stage trains a SGG model on a subset of training data including only annotated implicit relations, which allows the model to learn rich correlations in the data and encourages it to predict more informative implicit labels in the next stage.
The second stage includes an alternating procedure that imputes missing implicit labels on the subset of samples annotated with explicit relations, followed by training on both the annotated and imputed labels, called \emph{label refinement}.
In this stage, a model is prone to confirm to its own (wrong) predictions to achieve a lower loss as observed in semi-supervised learning (\eg~\cite{tarvainen2017mean, arazo2020pseudo}).
To prevent such overfitting, we regularize the model by a latent space augmentation strategy.
We demonstrate that our method yields significant performance gains in the SGG task for the standard and zero-shot settings on the Visual Genome~\cite{krishna2017visual} when applied to several existing scene graph generation models. 

In short, our contributions are as following. We identify a previously unexplored issue, \emph{missing informative labels} in the standard SGG benchmark and address this through a model agnostic training procedure based on alternating label imputation and model training with effective regularization strategies. This method can be incorporated into state-of-the-art SGG models and boosts their performance by a significant margin.

\section{Related work}
\label{sec.rel}




\noindent
\textbf{Scene Graph Generation. }SGG has been extensively studied in the past few years with the goal of better understanding the object relations in an image by either focusing on architecture designs \cite{li2017scene, xu2017scene, zellers2018neural, tang2018learning, yang2018graph, chen2019knowledge} or feature fusion methods \cite{zareian2020bridging, zareian2020weakly, li2018factorizable, zhang2019graphical,hung2019union, yan2020pcpl, dornadula2019visual, gu2019scene}.
Recently, Tang \etal \cite{tang2018learning, tang2020unbiased} reported that the performance gains from these methods largely come from improved performance only on the head classes (frequently occurring relations) while the performance on most other relations is poor. They propose replacing the biased evaluation metric Recall@k with mean-Recall@k to assign equal importance to all labels.

The same authors \cite{tang2020unbiased} report that the bias in the data often drives SGG models to predict frequent labels and propose to use counterfactual causality \ie measure the difference in predictions between the original scene and a counterfactual one to remove the effect of context bias and focus on the main visual effects of the relation. Chiou~\etal~\cite{chiou2021recovering} address the bias in scene graph generation by learning from positive and unlabeled object pairs. 
Suhail~\etal~\cite{suhail2021energy} propose an energy based loss that learns the joint likelihood of object and relations instead of learning them individually.
This helps to incorporate commonsense structure (\textit{man \textbf{riding} horse} and \textit{man \textbf{feeding} horse} occurring together are highly improbable) and in better context aggregation. 
Unlike~\cite{tang2020unbiased,suhail2021energy}, our focus is on extracting more information from the training data through mining informative labels.
In fact, we show that our model is orthogonal to theirs and boosts performance when incorporated to theirs.
The most similar to ours, Wang~\etal~\cite{wang2020tackling} propose a semi-supervised method that employs two deep networks, where the auxiliary one imputes missing labels of unlabeled pairs and self-trains on them and transfers its knowledge to the main network. 
Unlike \cite{wang2020tackling}, who treat all the labels equally, our method only imputes informative implicit labels.
This is crucial, as shown in \cref{table:ablation_impute}, because, without such consideration, imputing labels cannot extract any substantial information from unlabeled samples leading to only minor gains.
In addition, our framework is more efficient as it involves only a single network that jointly infers labels and trains on them, outperforming \cite{wang2020tackling} significantly.


\noindent
\textbf{Label Completion. }There is a rich body of work in the literature that focuses on learning from partial/missing labels in a multi-label learning setting where each image is labelled for multiple categories with some missing labels \cite{bucak2011multi, durand2019learning, mahajan2018exploring, cole2021multi, cabral2011matrix, huang2021multi}. 
Common strategies to address this can be divided into two categories: 1) graph based methods \cite{wu2018multi, huynh2020interactive} that exploit similarity between samples to predict missing labels, and 2) low rank matrix completion which extracts label correlations \cite{cabral2011matrix, durand2019learning, yang2016improving,xu2014learning} to complete missing labels.  
There is another setting in which some instances miss all the labels, also called semi-supervised learning in multi-label classification \cite{tan2017semi, zhao2015semi}. In this setting, the classifier is trained for unseen data.
While related to our setting, we look at the classification of relations conditioned on subject--object pairs (rather than on the image level), with each pair already labeled with one relation (rather than unlabeled images).
Finally, we group the label set in two groups and treat them asymmetrically in our training.



\noindent
\textbf{Semi-supervised learning.}
Semi-supervised learning methods exploit unlabeled data via either pseudo-labeling or imputation with small amounts of labeled data \cite{rizve2021defense, shi2018transductive} or by enforcing consistency regularization on the unlabeled data to produce consistent predictions over various perturbations of the same input \cite{tarvainen2017mean, verma2021interpolation} by applying several augmentation strategies such as Mixup \cite{zhang2017mixup}, RandAugment \cite{cubuk2020randaugment}, AutoAugment \cite{cubuk2018autoaugment} or combine both pseudo-labeling and consistency regularization \cite{berthelot2019mixmatch, sohn2020fixmatch}.
Inspired from pseudo-labeling in semi-supervised learning, the main motivation of our method is to impute informative missing labels to improve generalization and learn complex features by relying on partially labeled data and still predict more accurate labels on the biased test set.


\section{Methodology}
\label{sec.method}

\subsection{Revisiting SGG Pipeline}
In SGG, we seek to localize and classify the objects in an image followed by labeling the visual relations between each pair of objects (or subject and object). 
Concretely, let $C$ and $P$ denote the object and relation classes respectively. 
Each subject or object $e = (e^b, e^c) \in \mathcal{E}$ consists of a bounding box $e^b \in \mathbb{R}^4$ and a class label $e^c \in C$. 
A relation tuple is a triplet of the form $r = (s, p, o)$ where the subject $s$ and the object $o$ ($s, o \in \mathcal{E}$) are joined by the relation $p \in P$, \eg \emph{man \textbf{wearing} shirt}. 
Given an image $I$, we can then use a set of objects $E = {\{e_i\}}_{i=1}^m$ and a set of relations $R = {\{r_j\}}_{j=1}^n$, where $m$ and $n$ are the number of subject/objects and relation triplets in an image respectively, to define a scene graph $S = (E, R)$. 
A scene graph can also be written as a combination of a set of bounding boxes $B = {\{e_i^b\}}_{{i=1}}^m$, a set of class labels $Y = {\{e_i^c\}}_{{i=1}}^m$ and a set of relations $R$. 

The SGG models can be decomposed as:
\begin{equation}
    P(S|I) = P(B|I)~P(Y|B,I)~P(R|B,Y,I)
    \label{eq.probsgg}
\end{equation}
where $P(B|I)$ is the object detector or bounding box prediction model, $P(Y|B,I)$ is an object class model and $P(R|B,Y,I)$ is a relation prediction model. 

Existing methods \cite{lu2016visual, li2017vip, xu2017scene, zhuang2017towards, zhang2017visual, hung2020contextual, zellers2018neural, tang2018learning} often employ a two-step process for the scene graph generation task. 
First, bounding-box proposals ($P(B|I)$) with class predictions and confidence scores ($P(Y|B,I)$) are extracted using off-the-shelf object detectors \cite{ren2015faster, girshick2015fast}.
Then, a multimodal feature fusion model combines visual, language and spatial features to predict the relation for a given subject-object pair ($P(R|B,Y,I)$). 
Several methods adopt BiLSTMs \cite{zellers2018neural}, Bi-TreeLSTMs \cite{tang2018learning} or fully connected layers \cite{zhang2017visual, hung2020contextual} to encode the co-occurrence between object pairs for relation prediction. 

\subsection{Missing Relation Labels}
\label{sec.missing}

Many visual relations are hypernyms, hyponyms, or synonyms \cite{ramanathan2015learning, yang2021probabilistic} and hence are non-mutually exclusive. The standard SGG datasets (\eg Visual Genome \cite{krishna2017visual}) ignore this fact and only assume \emph{one} annotated label per subject--object pair. Which one is assumed strongly depends on the annotator (manifesting as labeling or language/reporting bias \cite{misra2016seeing}). 

One way to circumvent this problem is to collect multiple labels for each triplet, which is however expensive and time consuming. 
Another potential solution is to use linguistic sources such as WordNet \cite{miller1995wordnet} or VerbNet \cite{schuler2005verbnet} to automatically obtain the missing labels by exploiting the linguistic dependencies between relations.
However, this is not trivial, as some of the relation and spatial vocabulary in the SGG datasets are not included in WordNet. Moreover, the context of relations in these language resources does not always allow the right inferences. For instance, in WordNet, \emph{person \textbf{riding} horse} does not imply \emph{person \textbf{on} horse} (no hyponymy relation), but this is the visual implication in the SGG datasets. 

While one can use existing methods~\cite{wang2020tackling, chiou2021recovering} to infer the missing labels, the estimated labels can be noisy and uninformative such that re-training a model on them may not improve the generalization performance.
Here, inspired by previous work \cite{collell2018acquiring}, we propose to group visual relations into two sets: explicit and implicit.\footnote{Further details about the explicit and implicit relations is in the experiments and the supplementary.}
Explicit relations encode spatial information between two objects such as `on', `in front of' or `under' and are typically easy to learn, even only based on subject--object locations \cite{collell2018acquiring}.
The implicit ones are normally verbs such as `riding', `walking' and for learning them the model has to find complex correlations in visual and textual data.
In existing SGG datasets, some object pairs are annotated with implicit labels, while other pairs are labeled only with explicit ones and their implicit labels are missing.
We propose to divide the set of predicate labels $P$ into two sets, \textit{explicit} and \textit{implicit} and denote them by $\mathbb{E}$ and $\mathbb{I}$ respectively.


\subsection{Proposed Method} 

In this section, we explain our proposed method for training the relation classifier $f_{\theta}$ to implement $P(R|B,Y,I)$.
For each image $I$, we assume that an object detector provides a set of candidate subject--object pairs $\{(s-o)\}$ and each pair is represented by a $d$-dimensional joint embedding $\xb \in \mathbb{R}^d$ including its visual, semantic and spatial features.
Note that we apply our method to various existing SGG models which uses different object detector and joint embedding functions, and we provide these details in \cref{sec.exp}.
In particular, $f_{\theta}$ is instantiated as a deep neural network parameterized by $\theta$, takes in a joint feature embedding $\xb$ for a subject--object pair $s-o$ and outputs a softmax probability over $|P|$ relations, \ie $f_{\theta}(\xb): \mathbb{R}^d \rightarrow \mathbb{R}^{|P|}$.
Our goal is to learn a relation classifier $f_{\theta}$ that can correctly estimate the relation label of a subject-object pair in an unseen image.

Given a training set $\mathcal{D}$ with $|\mathcal{D}|$ samples, each including tuples of subject-object pairs $s-o$ along with the relation label $p$ and the joint feature embedding $\xb$, which we denote with $X=\{(s,p,o,\xb)\}_{i=1}^{|X|}$ with $|X|$ tuples.
We formulate the learning problem as minimization of two loss terms:
\begin{equation}
    \label{eq.bigobj}
    \min_{\theta} \frac{1}{|\mathcal{D}|} \sum_{i=1}^{|\mathcal{D}|} \big(\LLI(X_i;\theta) + \LLE(X_i;\theta) \big)
\end{equation} where $\LLI(X_i;\theta)$ and $\LLE(X_i;\theta)$ are the loss terms defined over implicit and explicit relations respectively.

For a given image $I$ and its tuple $X$, we pick the tuples whose relation is annotated only with implicit relation label (\ie $X^{\mathbb{I}} = \{ (s,p,o,\xb) ~|~ p \in \mathbb{I}\}$ ) and define the implicit loss term as:
\begin{equation}
    \label{eq.impobj}
    \LLI(X^\mathbb{I};\theta) =  \frac{1}{|X^{\mathbb{I}}|} \sum_{(s,p,o,\xb) \in X^\mathbb{I}} \CE ( f_{\theta}(\xb),p)
\end{equation} where $\CE$ is the cross-entropy loss.
In other words, for the implicit relations, we follow the standard practice and compute its loss by using its ground-truth implicit relation  label $p$ which is an one-hot vector, as each subject--object pair is annotated with only one label.

Similarly, we formulate the explicit term $\LLE(X^\mathbb{E};\theta)$ for the tuples with explicit labels:
\begin{equation}
    \label{eq.expobj}
    \LLE(X^\mathbb{E};\theta) =  \frac{1}{|X^{\mathbb{E}}|} \sum_{(s,p,o,\xb) \in X^\mathbb{E}} \KL ( f_{\theta}(\xb),\hat{p})
\end{equation} where $\KL$ is the Kullback-Leibler divergence, $X^{\mathbb{E}} = \{ (s,p,o,\xb) ~|~ p \in \mathbb{E}\}$  and $\hat{p} $ is the \emph{imputed} relation label for the subject-object pair, which is a vector with soft probabilities.
Next, we discuss our method to obtain $\hat{p}$.

\begin{figure}[t]
\begin{center}
\includegraphics[width=\linewidth]{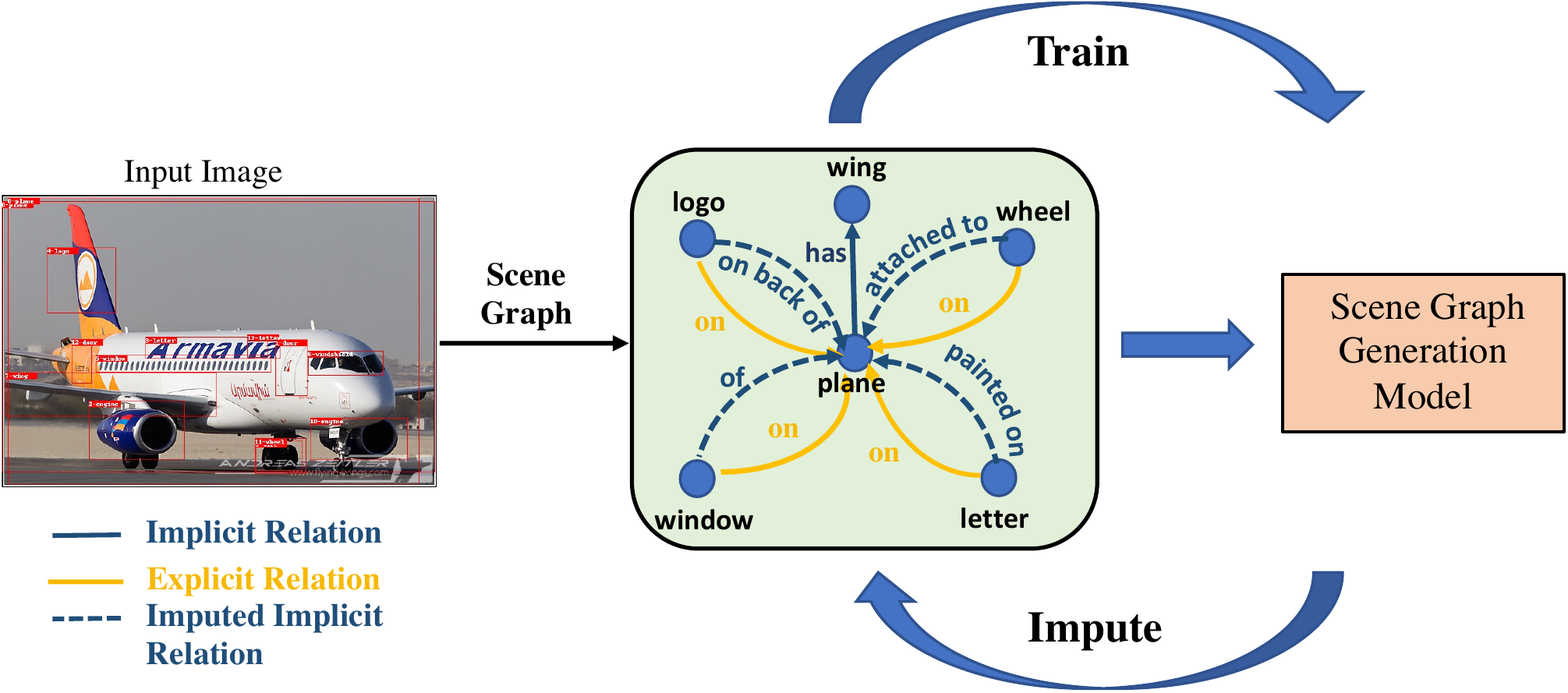}
\vspace{-.8cm}
\end{center}
\caption{Pipeline of our proposed framework for mining informative labels for scene graph generation. An input image can be represented as a scene graph (in green). The yellow and blue solid arrows in the scene graph denote ground-truth explicit and implicit relations, respectively. The scene graph generation model is first trained on a subset of the implicit relations. In an iterative fashion, we then impute implicit relations (dashed blue arrows) for the triplets annotated with explicit relations and train the model with all relations (implicit, explicit, and imputed).}
\label{fig:high-level-idea}
\end{figure}

\noindent
\textbf{Label imputation.}
The subject--object pairs that are annotated with explicit relations only can often be labelled with more informative implicit relation labels.
For instance, the ground truth label may be \textit{person \textbf{beside} table}, but \textit{person \textbf{eating at} table} might also be correct in this case, and more informative. 
To impute the missing implicit relation for a subject--object pair, which is originally annotated with an explicit relation label, we follow a two-step procedure.

First, we take each subject--object pair annotated with an explicit label from $X^{\mathbb{E}}$ along with its joint embedding and impute its implicit label through the relation classifier $f_\theta$ as:
\begin{equation}
    \bar{p} = \argmax_{i \in \mathbb{I}} \sbr{\frac{\text{exp}(f_{\theta}^i(\xb))} { \sum_{j \in \mathbb{I}}~\text{exp}(f_{\theta}^j(\xb))}}
    \label{eq.impute}
\end{equation} where $f^i$ denotes its logit for the $i$-th relation class.
In words, we compute softmax probabilities only over implicit relation labels and pick the highest scoring implicit label to obtain an one-hot vector $\bar{p}$.
Note that one can also obtain a soft probability vector over all implicit label classes, however, we empirically show that the former works better.

Second, as the subject--object pair is originally labelled with an explicit label $p$, we also use this information and average the imputed label $\bar{p}$ with its original label $p$ as:
\begin{equation}
    \label{eq.phat}
    \hat{p} =  (p + \bar{p}) / 2.
\end{equation}
We call this step as \textbf{\emph{Label Refinement}}. 
As $\hat{p}$ includes equal probabilities (\ie 0.5) for each of the explicit and implicit label, it is not an one-hot vector.
Thus, we use KL divergence \cref{eq.expobj} to encourage the model to predict both the labels. Compared to standard cross-entropy loss, the KL divergence
loss increases the model entropy by reducing overconfidence, resulting in smoother predictions. Most of the traditional methods for SGG trained with cross-entropy may get confused by inconsistent annotations, where same relation is labeled with less informative spatial relation in some
images while more informative labels are used in some
other images. Our loss function formulation and multi-label
nature of the targets addresses this inconsistency, unlike
previous work \cite{tang2018learning, suhail2021energy}. 

\noindent
\textbf{Latent space augmentation.}
Our training pipeline is illustrated in \cref{fig:high-level-idea} which follows an alternating optimization and consists of two alternating steps where we employ the relation classifier $f_\theta$ to impute the implicit labels and simultaneously optimize the classifier parameters $\theta$.
The main challenge here is that the model parameters can overfit to its own imputed labels quickly, resulting in a local optimum solution.
This problem is notoriously known as confirmation bias that also occurs in many semi-supervised problems \cite{arazo2020pseudo, sohn2020fixmatch}.
To prevent overfitting to the wrong imputed labels, existing solutions include applying various kinds of data augmentation including standard geometric and color transformations \cite{devries2017dataset}, their combinations \cite{cubuk2018autoaugment, yun2019cutmix, cubuk2020randaugment} and also generating augmented version of samples \cite{zhang2017mixup,verma2019manifold}.

As many SGG methods build on the feature space of an object detector, many augmentation strategies that are applied on raw pixels are not applicable to our case.
Hence, we use Manifold Mixup \cite{verma2019manifold} that generates augmented embeddings (in the maninfold space rather than in the pixel space) by taking a convex combination of different pairs of embeddings ($\xb$ and $\xb'$) and also their labels ($p$ and $p'$):
\begin{equation}
    \label{eq.mmixup}
    \begin{split}
        \tilde{\xb} = \lambda . \xb + (1 - \lambda). \xb' \\
        \tilde{p} = \lambda . p + (1 - \lambda). p'
    \end{split}     
\end{equation}
where $\lambda$ is sampled from a beta distribution, \ie $\lambda \sim \text{Beta}(\alpha, \alpha)$ with $\alpha$ as a hyperparameter.
Note that we apply this augmentation to the whole training set and allow mixing embeddings across samples from both the implicit and explicit set of relations. This augmentation acts as a regularizer and accounts for overfitting to the incorrect imputed labels while training.

\subsection{Algorithm}
In \Cref{algo:vrd}, we detail our training pipeline. 
To obtain the initial parameters $\theta_0$, we first train our model on the tuples with only implicit labels as following (Line 2):
\begin{equation}
    \label{eq.init}
    \theta_0=\argmin_{\theta} \frac{1}{|\mathcal{D}|} \sum_{i=1}^{|\mathcal{D}|} \LLI(X_i;\theta).
\end{equation}
The key intuition behind this is that model learned on only the implicit relations are more likely to produce confident predictions over implicit labels and hence not `distracted' by explicit relation labels.

After the training on implicit relations, we iteratively impute implicit relation labels for the subject-object pairs annotated with the explicit relations (Line 5) and update the model parameters using \cref{eq.bigobj} (Line 7). 
The model parameters optimized in \cref{eq.bigobj} take as input the augmented versions of the sample and label pair (as in \cref{eq.mmixup}) for both the implicit and explicit set of relation labels (Line 6).


\begin{algorithm}[h]
    \caption{Our proposed optimization of the SGG model}
     \begin{algorithmic}[1]
         \State \textbf{Input:}  Training set $\mathcal{D}$ with $|\mathcal{D}|$ samples, each including a set of tuples $X$ with $(s, p, o, \xb)$ with subject, object and relation label, joint embedding resp., Explicit and implicit relation sets $\mathbb{E}$ and $\mathbb{I}$ resp., relation classifier $f_{\theta}$, $T$ number of iterations, $\eta$ learning rate.
         \State Initialize $\theta$ as in \cref{eq.init}
         \For{$t=0,\dots,T$}
         \State Sample a minibatch $B=(X_1,\dots,X_{|B|}) \sim \mathcal{D}$,
         \State \textbf{Impute $\hat{p}$}: Impute implicit labels $\hat{p}_t$ for $B^{\mathbb{E}}$ by using \cref{eq.impute} and \cref{eq.phat},
         \State Augment $B$ by applying manifold mixup in \cref{eq.mmixup},
         \State \textbf{Update $\theta$}: $\theta_{t+1} \leftarrow \theta_t + \eta \Delta_{\theta} $ where $\Delta_{\theta}$ is the update for $\theta$ obtained from \cref{eq.bigobj},
         \EndFor
         \State \textbf{return $\theta$} 
 
\end{algorithmic}
\label{algo:vrd}
\end{algorithm}

\section{Experiments}
\label{sec.exp}
\begin{table*}[!ht]
\renewcommand*{\arraystretch}{1.13}

		\resizebox{0.9\textwidth}{!}{
			\begin{tabular}{c|c |c | c| c| c| c| c| c| c| c}
				\hline
				& & \multicolumn{3}{c|}{Predicate Classification} &\multicolumn{3}{c|}{Scene Graph Classification} & \multicolumn{3}{c}{Scene Graph Detection }\\
				
				 Models & Method & mR@20 & mR@50 & mR@100 & mR@20 & mR@50 & mR@100  & mR@20 & mR@50 & mR@100  \\
				\hline
				IMP \cite{xu2017scene, chen2019knowledge} & - & - & 9.8 & 10.5& - & 5.8 & 6.0 & - &3.8& 4.8\\
				FREQ \cite{zellers2018neural, tang2018learning}& - & 8.3 & 13.0 & 16.0 & 5.1 & 7.2 & 8.5 & 4.5& 6.1 & 7.1 \\
				Motifs \cite{zellers2018neural} & - & 10.8 & 14.0 & 15.3 & 6.3 & 7.7 & 8.2& 4.2& 5.7& 6.6 \\
				KERN \cite{chen2019knowledge} & - & - & 17.7 & 19.2 & - & 9.4 & 10.0&  - &6.4 &7.3 \\
				VCTree \cite{tang2018learning} & - & 14.0 & 17.9 & 19.4 & 8.2& 10.1 &10.8 &5.2 &6.9& 8.0\\ 
				VCTree-L2+cKD \cite{wang2020tackling} & - & 14.4 & 18.4 & 20.0 & 9.7 & 12.1 &13.1 &5.7 &7.7& 9.0\\ \hline

 				\multirow{2}{*}{IMP \cite{xu2017scene}} & Baseline &  8.9 & 11.0 & 11.8 & 5.4 & 6.4& 6.7 &2.2 &3.3& 4.1 \\ 
				& Ours & \textbf{12.3} & \textbf{14.6} & \textbf{15.3} & \textbf{7.1} & \textbf{8.0} & \textbf{8.3} & \textbf{6.9} & \textbf{7.8} & \textbf{8.1}  \\\hline

				
    			\multirow{2}{*}{Motif-TDE-Sum \cite{zellers2018neural, tang2020unbiased}} & Baseline & 17.9 & 24.8 & 28.7 & 9.8 & 13.2 & 15.1 & 6.6 & 8.9 & 11.0 \\ 
				 & Ours  & \textbf{21.3} & \textbf{27.1} & \textbf{29.7} & \textbf{11.3} & \textbf{14.3} & \textbf{15.7}  &\textbf{8.4} & \textbf{10.4} & \textbf{11.8} \\ \hline

				\multirow{2}{*}{VCTree \cite{tang2018learning}} & Baseline & 13.1 & 16.5 & 17.8 & 8.5&  10.5 & 11.2 & 5.3 & 7.2 & 8.4 \\ 
			
				 & Ours & \textbf{18.0} & \textbf{21.7} & \textbf{23.1} & \textbf{11.9} & \textbf{14.1} &  \textbf{15.2} & \textbf{7.1}  & \textbf{8.2} & \textbf{8.7} \\ \hline
				
				\multirow{2}{*}{VCTree-EBM \cite{suhail2021energy}} & Baseline &14.2&  18.2& 19.7& 10.4 &12.5& 13.4& 5.7 &7.7 &9.1   \\ 
			
				 & Ours & \textbf{21.0} & \textbf{24.9} & \textbf{26.5} & \textbf{14.0} & \textbf{16.2} & \textbf{17.1} & \textbf{7.8} & \textbf{10.1} & \textbf{11.8} \\ \hline
				
				\multirow{2}{*}{VCTree-TDE \cite{tang2020unbiased}} & Baseline & 16.3 & 22.9 & 26.3 & 11.9 & 15.8 & 18.0 & 6.6 & 9.0 & 10.8 \\ 
				
                 & Ours & \textbf{22.2} & \textbf{28.1} & \textbf{30.6}  & \textbf{17.8} & \textbf{22.0} & \textbf{23.6} & \textbf{8.4} & \textbf{10.3} & \textbf{11.5} \\ \hline

\end{tabular}}
\caption{Scene Graph Generation performance comparison on mean Recall@K \cite{tang2020unbiased} under all three experimental settings. We compare the results of our proposed framework (Ours) with the original model (Baseline) using different SGG architectures.}
\label{table:sota_mr}
\end{table*}

\noindent
\textbf{Dataset and Evaluation Settings. }
We evaluate our proposed method for scene graph generation on the Visual Genome (VG) \cite{krishna2017visual} dataset. We use the pre-processed version of the VG dataset as proposed in \cite{xu2017scene}. The datatset consists of 108k images with 150 object categories and 50 relation categories. The training, test and validation split used in the experiments also follow previous work \cite{xu2017scene, tang2020unbiased, suhail2021energy}. 

For evaluation on the \textbf{Visual Genome dataset}, we follow \cite{xu2017scene} and report performance on three settings:  \textbf{(1)~Predicate Classification (PredCls).} This task measures the accuracy of relation (also termed as predicate in literature) prediction when the ground truth object classes and boxes are given. It is not affected by the object detector accuracy. \textbf{(2)~Scene Graph Classification (SGCls).} In this setting, we know the ground truth boxes and we have to predict the object classes and the relations between them. \textbf{(3)~Scene Graph Detection (SGDet).} This is the most challenging setting and the models are used to predict object bounding boxes, object classes and the relations between them. We measure Mean Recall@K (mR@K) \cite{tang2018learning, tang2020unbiased} to evaluate scene graph generation models. More recent work has preferred mR@K over regular Recall@K \cite{xu2017scene} due to data imbalance \cite{tang2020unbiased}.  Mean Recall@K treats each relation separately and then averages Recall@K over all relations. 
We also measure the \textbf{zero-Shot Recall}, zsR@K, for three settings of PredCls, SGCls and SGDet, which helps to evaluate the generalization ability of the model in predicting subject--relation--object triplets not seen during training.




\noindent
\textbf{Model Generalization. }
Our proposed framework has the flexibility to be trained with any scene graph generation model. Hence, we train with different model architectures to demonstrate the generalizability of our approach: Iterative Message Passing (IMP) \cite{xu2017scene}, Neural-Motifs \cite{zellers2018neural} and VCTree \cite{tang2018learning}. We also train with two other debiasing methods that build upon these models, Energy-based Modeling (EBM) \cite{suhail2021energy}, where the authors propose to train with an additional energy-based loss and Total Direct Effect (TDE). where counterfactual reasoning is used during inference. 

\noindent
\textbf{Explicit and Implicit Relation Labels. }
For the Visual Genome dataset \cite{krishna2017visual}, Xu \etal \cite{xu2017scene} released a version of the dataset with 50 relations which are: \emph{above, across, against, along, and, at,  attached to,  behind,  belonging to,  between,  carrying,  covered in, covering, eating,  flying in,  for,  from,  growing on,  hanging from,  has,  holding,  in,  in front of, laying on, looking at,  lying on,  made of,  mounted on,  near,  of,  on,  on back of,  over,  painted on,  parked on, part of,  playing,  riding,  says,  sitting on,  standing on,  to,  under,  using,  walking in,  walking on,  watching,  wearing,  wears,  with.}

Inspired by \cite{collell2018acquiring}, we divide the relation label space into a set of explicit and implicit relations. We define explicit relations when the spatial arrangement of objects are implied by the label itself \eg,  ``on",  ``below", ``next to'' and so on.   For implicit relations,  the spatial arrangement of objects is only indirectly implied, ``riding'', ``walking'', ``holding" etc. More specifically, the explicit relations are \emph{above, across, against, along, at, behind, between, in, in front of, near, on, over, under} and the rest are treated as implicit relations.

\noindent
\textbf{Implementation Details. }
Following previous work \cite{tang2018learning, tang2020unbiased}, we train the scene graph generation models on top of the pre-trained Faster R-CNN object detector with ResNetXt-101-FPN backbone \cite{ren2015faster}. The weights of the SGG models' object detector are frozen during training in all the three settings -- PredCls, SGCls and SGDet. The mAP of the object detector on the Visual Genome dataset is 28\% using 0.5 IoU. For training each scene graph generation model using our proposed method, we use the default training settings as in \cite{tang2020unbiased, suhail2021energy} for fair comparisons. The models are trained with the SGD optimizer with a batch size of 12, an initial learning rate of $10^{-2}$ and 0.9 momentum. 

The models are trained for the first 30,000 batch iterations on the implicit label subset with the standard cross-entropy loss. After label imputation, the model is trained on the rest of the imputed data and the implicit subset for another 20,000 batch iterations. The value of $\alpha$ is set to 4 from which $\lambda$ is sampled for the mixing function in the latter half of training. 
Our code is available here\footnote{ \url{https://groups.inf.ed.ac.uk/vico/research/NARE}}. 

\begin{table}[t]

		\resizebox{0.95\textwidth}{!}{
			\begin{tabular}{c|c |c | c| c}
				\hline
				& & \multicolumn{1}{c|}{PredCls} &\multicolumn{1}{c|}{SGCls} & \multicolumn{1}{c}{SGDet}\\
				
				 Models & Method & zsR@20/50 & zsR@20/50  & zsR@20/50  \\
				\hline

				\multirow{2}{*}{IMP \cite{xu2017scene}} & Baseline &  \textbf{12.17/17.66} & \textbf{2.09/3.3} & 0.14/0.39 \\ 
				
				& Ours & 7.12/10.50 & 1.57/2.32 & \textbf{1.52/2.48} \\\hline
		
				
    			\multirow{2}{*}{Motif-TDE-Sum \cite{tang2020unbiased}} & Baseline & 8.28/14.31 & \textbf{1.91}/2.95& 1.54/2.33 \\ 
				
				& Ours  & \textbf{9.33}/\textbf{14.43} & 1.87/\textbf{2.99} & \textbf{2.06}/\textbf{3.05} \\ \hline

				\multirow{2}{*}{VCTree \cite{tang2018learning}} & Baseline &  1.43/\textbf{4.0} & \textbf{0.39/1.2} & 0.19/0.46 \\ 
			      & Ours & \textbf{1.51}/3.7  & 0.36/1.0 & \textbf{0.43/0.95} \\ \hline  

			

				\multirow{2}{*}{VCTree-TDE \cite{tang2020unbiased}} & Baseline &  8.98/\textbf{14.52} & 3.16/4.97 & 1.47/2.3 \\ 
				& Ours &\textbf{9.11}/13.52 & \textbf{4.26}/\textbf{6.20} & \textbf{2.24}/\textbf{3.25} \\ \hline

\end{tabular}}
\caption{Zero shot recall performance for our proposed method compared with the original model (baseline).}
\label{table:sota_zsr}
\end{table}

\section{Results}
\noindent
\textbf{Quantitative Results.} \Cref{table:sota_mr} compares the performance of the state-of-the-art methods when trained with our proposed training framework on the Visual Genome dataset \cite{krishna2017visual}. Our method consistently improves performance on all the three evaluation settings (PredCls, SGCls, SGDet) when trained with existing methods. With IMP \cite{xu2017scene} and Motif-TDE-Sum \cite{zellers2018neural, tang2020unbiased}, we obtain an absolute improvement of 3.4\% and 1.7\% in PredCls and SGCls, respectively. For the most challenging setting of SGDet, there is an improvement of 4.7\% with the IMP model, showing the generalization ability of our approach on a visual-only model (no language/textual features). 
When debiasing approaches such as TDE \cite{tang2020unbiased} and EBM \cite{suhail2021energy} are incorporated, we obtain consistent improvements for VCTree.
In both the cases, we gain significantly in all the settings and achieve a new state-of-the-art performance in scene graph generation by combining our proposed method with VCTree-TDE.  

In \Cref{table:sota_zsr}, we report the zero-shot recall performance and compare it with baselines. Our proposed method achieves improvements in most of the settings with different SGG backbones, except for IMP in PredCls and SGCls. IMP being a visual-only model fails to learn correlations via textual features for different relation classes and hence performs poorly in zero-shot, due to low recall on explicit relations. However, the multi-modal nature of Motif and VCTree brings out the strength of our method in generalizing to unseen triplets during test time.

\begin{table}[!ht]
\renewcommand*{\arraystretch}{1.13}

		\resizebox{0.8\textwidth}{!}{
			\begin{tabular}{c |c | c| c|c}
				\hline
				&  \multicolumn{4}{c}{Predicate Classification} \\
				
				 Method & Train Label & Imputed with & Imputed on & mR@20/50 \\
				\hline

				 \multirow{4}{*}{Baseline} & All & - & - & 17.85/24.75 \\
				 
				 & Explicit & - & - & 14.06/20.34 \\
				 & Random & - & - & 16.99/23.33 \\
				 & Implicit & - & - & 18.24/24.93 \\ \hline
    			 \multirow{3}{*}{Ours} & Random & Top1 & Random & 17.11/23.56\\ 
    			 
    			  & Explicit & Top1-Explicit & Implicit & 14.23/20.51 \\  
    			  &  Implicit & Top1-Implicit & Explicit & \textbf{21.26/27.14}  \\ \hline

\end{tabular}}
\caption{Experimental results on the Predicate Classification setting with different ways of label imputation.}
\label{table:ablation_impute}
\end{table}

\begin{figure*}
\begin{center}
\includegraphics[width=0.78\linewidth]{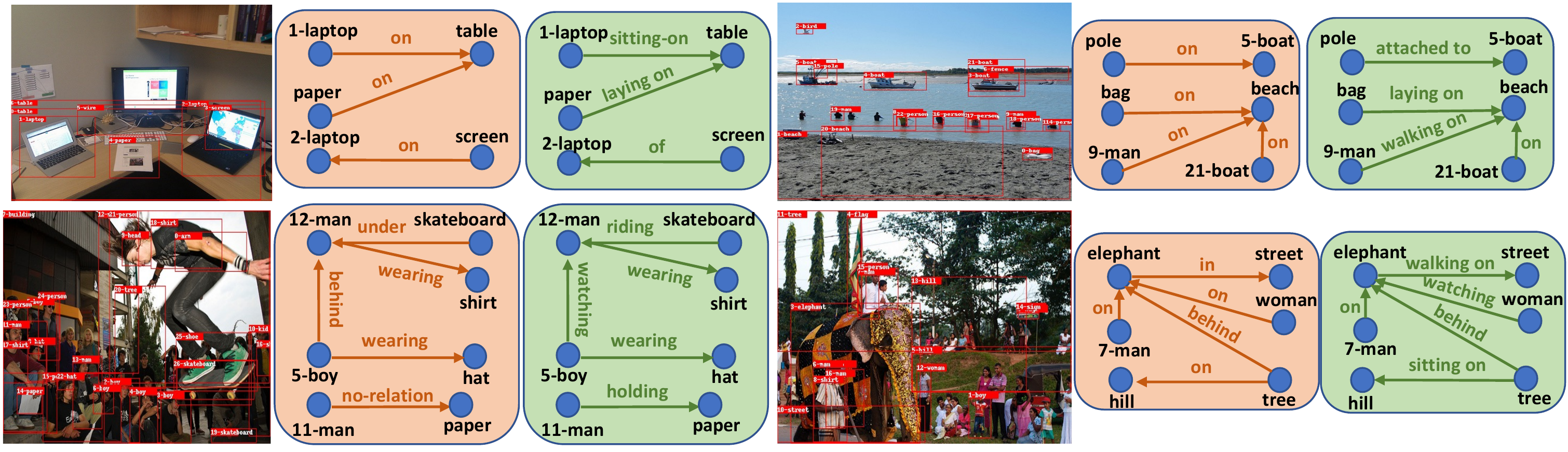}
\vspace{-.5cm}
\end{center}
\caption{Visualization of scene graphs generated by the VCTree-EBM \cite{suhail2021energy} based learning framework (in orange) and our proposed method using VCTree-EBM as backbone (in green). }
\label{fig:qualresultsfinal}
\end{figure*}

\begin{figure*}
\begin{center}
\includegraphics[width=0.81\linewidth]{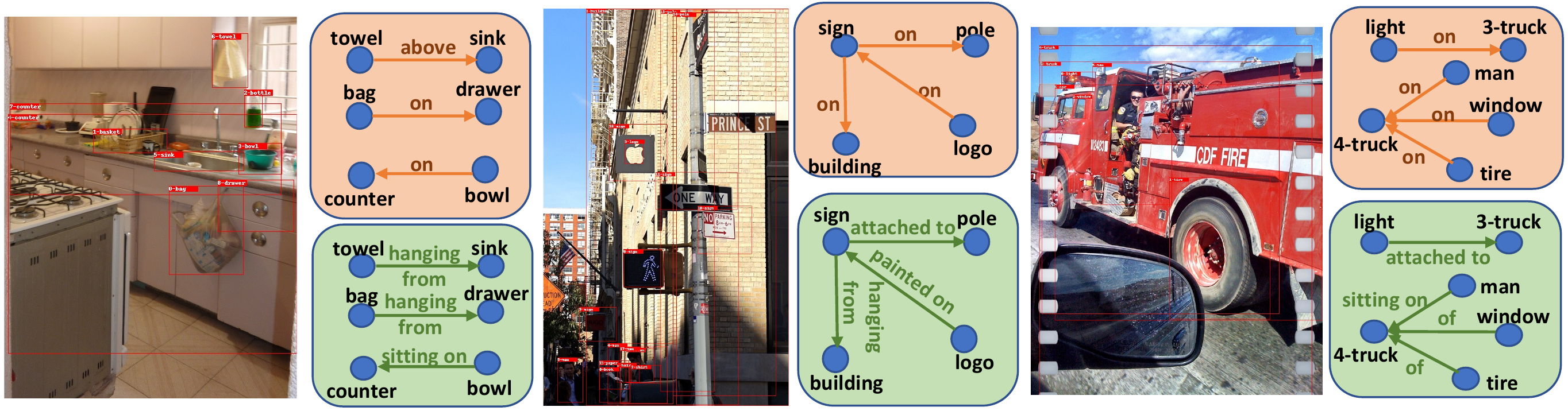}
\end{center}
\caption{Scene Graph Visualization results for the missing \textit{implicit} relations imputed for the triplets annotated with \textit{explicit} relations. The scene graphs in orange are the ground-truth triplets with their corresponding imputed \textit{implicit} relations in green. }
\label{fig:qualresultsimpute}
\vspace{-0.5cm}
\end{figure*}

\noindent
\textbf{Ablative Analysis.} In this section, we study different components of our method separately to validate their effectiveness. \Cref{table:ablation_impute} evaluates the contribution of our proposed label imputation strategy. All the models are trained using the Motif-Sum \cite{zellers2018neural} backbone with TDE \cite{tang2020unbiased} at inference time, Motif-TDE-Sum. \textit{Random} implies that we randomly divide the relation labels into two sets without the knowledge of \textit{explicit} and \textit{implicit} relations. The results in the rows for the method \textit{Baseline} show the effect of training with \textit{implicit} relations on scene graph generation performance. If the model is trained with the \textit{explicit} relations only, mean recall drops. There is also a marginal drop in performance compared to training with all relations when trained on a random subset of the relations. Even training on a subset of the data with only implicit relations, we can achieve at-par performance compared to training with all relations. 

We also compare the results under settings when the model is trained on a particular subset of \textit{Train Label} with different imputation strategies (results in rows with the method \textit{Ours}). The relation labels are imputed on the hold-out relation set with the top-1 (best) in the set of labels that it was trained on. Training and imputing with the random or explicit set of labels decreases performance compared to the baseline, with a significant drop when only explicit relations are considered. This shows the importance of learning with and imputing implicit relations: they provide useful information about interactions between object pairs not captured by explicit relations. 


				

				 


\begin{table}[!ht]

		\resizebox{0.8\linewidth}{!}{
			\begin{tabular}{c |c | c| c | c}
				\hline
				&  \multicolumn{4}{c}{PredCls} \\
				
				  Method & Mixup & Soft/Hard Labels & Refinement &mR@20/50  \\
				\hline

				\multirow{2}{*}{Baseline}& \tickno &\tickno& \tickno & 17.85/24.75 \\ 
    			 & \tickyes & \tickno & \tickno & 17.43/24.20  \\ \hline
    			 
    			 \multirow{5}{*}{Ours}& \tickno & Hard & \tickno & 18.26/24.23 \\
    			  & \tickyes & \tickno & \tickyes & 18.90/25.32 \\

    			 & \tickno & Hard & \tickyes & 20.81/26.78  \\
    			 & \tickyes & Hard & \tickno & 19.90/26.35 \\
    			 & \tickyes & Soft & \tickyes & 20.76/27.10 \\ \cline{2-5}
				& \tickyes & Hard & \tickyes & \textbf{21.26/27.14} \\ \hline

\end{tabular}}
\caption{Ablation study on our proposed training framework with Motif-TDE-Sum \cite{zellers2018neural, tang2020unbiased} as the SGG backbone.}
\label{table:ablation_loss}
\end{table}

In \Cref{table:ablation_loss}, we study whether our method improves over the baselines mostly due to training with \textit{Latent Space Augmentation} (Manifold Mixup) or combination of multiple components. Applying manifold mixup on either the baseline (row 2) or with only explicit ground-truth labels without imputation (row 4),  does not provide significant gains. This indicates that mixup is effective only when applied to the imputed implicit labels. This hints that mixup can reduce overfitting to noisy imputations and allow for imputing more informative labels. We also show that the performance of soft imputed labels (row 7) is very similar to our proposed method (with hard labels, row 8). Using hard labels reduces noise in the predictions and encourages the model to learn from more high-confident predictions.


In \Cref{table:expimp}, we show the results for training on relation subsets and our final method on the explicit and implicit set of relations separately. As discussed in \Cref{sec.intro}, when the model is only trained on explicit relations, it fails to generalize to implicit relations. This is in contrast to training only on the implicit set of relations.  This indicates that implicit relations are rich in information and perhaps learn complex and generalizable features. Our final method outperforms training on all relations (original model) and the subset of relations by a significant margin, showing the strength of mining  informative labels for less informative samples.

\begin{table}[!ht]
\renewcommand*{\arraystretch}{1.13}

		\resizebox{0.97\textwidth}{!}{
			\begin{tabular}{c |c | c| c}
				\hline
				&  \multicolumn{3}{c}{PredCls - mR@50/100} \\
				
				 Method & Train Relations $\downarrow$/Test Relations $\rightarrow$ & Explicit & Implicit  \\
				\hline

    			 \multirow{4}{*}{MOTIF-TDE-Sum} & All Relations& 24.47/28.79 & 24.96/28.74\\ 
    			 
    			  & Explicit only & 22.89/25.34 & 0.08/0.09  \\  
    			  &  Implicit only & 20.10/22.89 &  24.34/26.03  \\ \cline{2-4}
    			  & Ours (final) & \textbf{24.83/27.80} &  \textbf{27.99/30.38}  \\ \hline
			
\end{tabular}}
\caption{Performance Comparison on the Explicit and Implicit set of relations separately with different subsets of training labels.}
\label{table:expimp}
\end{table}

\noindent
\textbf{Qualitative Results.} \Cref{fig:qualresultsfinal} visualizes the scene graphs predicted from the baseline VCTree-EBM \cite{suhail2021energy} model (in orange) and compares it to the scene graphs obtained via our proposed training framework (in green). Our method consistently predicts more informative relations such as \textbf{\textit{laying on, walking on, holding}} instead of simple prepositional relations such as \textbf{\emph{on, in}}. Moreover, our method also effectively identifies triplets with relations that were missed in the baseline. For instance, in the bottom-left image, our method localizes \emph{man \textbf{holding} paper} correctly. Our method also corrects relations which are incorrectly predicted in the baseline, \emph{woman \textbf{watching} elephant} as opposed to \emph{woman \textbf{on} elephant} in the bottom-right image. 

In \Cref{fig:qualresultsimpute}, we visualize the imputed \textit{implicit} relations for the triplets annotated with explicit relations. In orange, we show the ground-truth scene graphs and the corresponding imputed scene graphs in green. It can be clearly observed from the ground-truth scene graphs that there is annotator bias towards spatial relations. Our label imputation strategy is able to find alternate and missing implicit relations for these triplets and exploit them during training. For instance, our method imputes important relations such as \textbf{\emph{attached to, hanging from, sitting on}} which are more descriptive than their explicit counterpart \textbf{\emph{on}}. This shows the importance of label imputation and generating descriptive scene graphs for comprehensive scene understanding.

\section{Conclusion}
\label{sec.conclusion}
We proposed a novel model-agnostic training framework for scene graph generation. We introduced the concept of label informativeness, which had not been explored in SGG before. A model trained on informative relations is able to model the visual and textual context better compared to training on simple spatial relations. We showed how to impute informative relations from the partially labeled data and jointly train with imputed and ground truth relations. We  improved the performance across models and tasks, including in a zero-shot setting. One limitation of our approach is its limited ability in predicting relations with very few samples, which should be investigated in future work.





{
\noindent
{\small
\textbf{Acknowledgement:} AG is supported by the Armeane Choksi Scholarship and HB is supported by the EPSRC programme grant Visual AI EP/T028572/1. This research/project is supported by the National Research Foundation, Singapore under its AI Singapore Programme  (AISG Award No: AISG2-RP-2020-016). We thank the anonymous reviewers for their constructive feedback.}}

{\small
\bibliographystyle{ieee_fullname}
\bibliography{refs}
}



\end{document}


\title{Supplementary Material -- Not All Relations are Equal:\\ Mining Informative Labels for Scene Graph Generation  }

\author{\parbox{16cm}{\centering
    
    {\large Arushi Goel$^1$, Basura Fernando$^{2}$, Frank Keller$^{1}$, and Hakan Bilen$^{1}$}\\
    {\normalsize
    $^1$School of Informatics, University of Edinburgh, UK \\
    $^2$CFAR, IHPC, A*STAR, Singapore.}}
}

\maketitle

\label{sec.supp}

\begin{strip}
\begin{center}
\includegraphics[width=\linewidth]{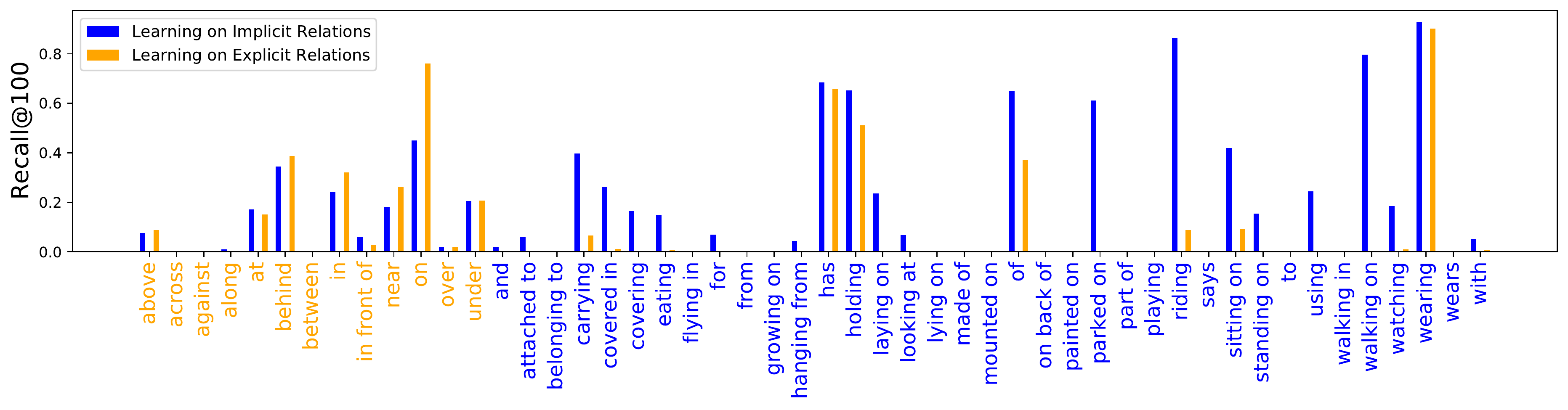}
\end{center}
\captionof{figure}{Class Wise Recall@100 for the set of Implicit and Explicit Relations by training on a subset of relations. All models are trained with Motif-TDE-Sum \cite{zellers2018neural, tang2020unbiased}.}
\label{fig:intro-fig-full}
\end{strip}

\begin{table*}[t]
    \begin{center}
		\centering
		\resizebox{\textwidth}{!}{
			\begin{tabular}{|c|c |c | c| c| c| c| c| c| c| c|c|c|c|}
				\hline
				 \textbf{Explicit Relations} & above&	across&	against&	along& 	at& 	behind&	between&in&	in front of	&near&	on&	over&	under \\
				\hline 
				  \# of Instances & 47341 &	1996 &	3092 &	3624 & 	9903 & 	41356 &	3411 &	251756 &13715	& 96589 &	712409 & 9317 &	22596 \\ \hline

\end{tabular}}
\end{center}
\caption{Explicit Relations for the Visual Genome Dataset \cite{krishna2017visual}.}
\label{table:expset}
\end{table*}

\begin{table*}[t]

		\resizebox{\textwidth}{!}{
			\begin{tabular}{|c|c |c | c| c| c| c| c| c| c| c|c|c|c|c|c |c | c| c| c| c| c| c| c| c|c|c|c|c|c |c | c| c| c| c| c| c| c|}
				\hline
				
				 \textbf{Implicit Relations} & attached to&	and&	belonging to&	carrying&	covered in &	covering&	eating	&flying in&	for& 	from	&growing on&	hanging from	&has&	holding	&laying on&	looking at \\
				\hline
			
				  \# of Instances & 10190 &	3477 &	3288&	5213&	2312 &	3806&	4688	&1973 &	9145& 	2945& 1853& 9894& 277936& 42722& 3739& 3083 \\ \hline
				
				\hline

\end{tabular}}
\caption{Implicit Relations for the Visual Genome Dataset \cite{krishna2017visual}.}
\label{table:impset1}
\end{table*}

\begin{table*}[!ht]

		\resizebox{\textwidth}{!}{
			\begin{tabular}{|c|c |c | c| c| c| c| c| c| c| c|c|c|c|c|c |c | c| c| c| c| c| c| c| c|c|c|c|c|c |c | c| c| c| c| c| c| c|}
				\hline
				
				 \textbf{Implicit Relations} & lying on&	made of& 	mounted on	&of	&on back of&	painted on&	parked on&	part of&	playing	&riding	&says&	sitting on& 	standing on&	to	&using	&walking in&	walking on&	watching&	wearing&	wears	&with \\
				\hline
			
				  \# of Instances & 1869& 2380& 2253&  146339. & 1914& 3095& 2721& 2065& 3810& 8856& 2241& 18643& 14185& 2517& 1925& 1740& 4613& 3490& 136099& 15457& 66425 \\ \hline
				
				\hline

\end{tabular}}
\caption{Implicit Relations for the Visual Genome Dataset \cite{krishna2017visual}.}
\label{table:impset2}
\end{table*}

\begin{figure*}[!h]
\begin{center}
\includegraphics[width=\linewidth]{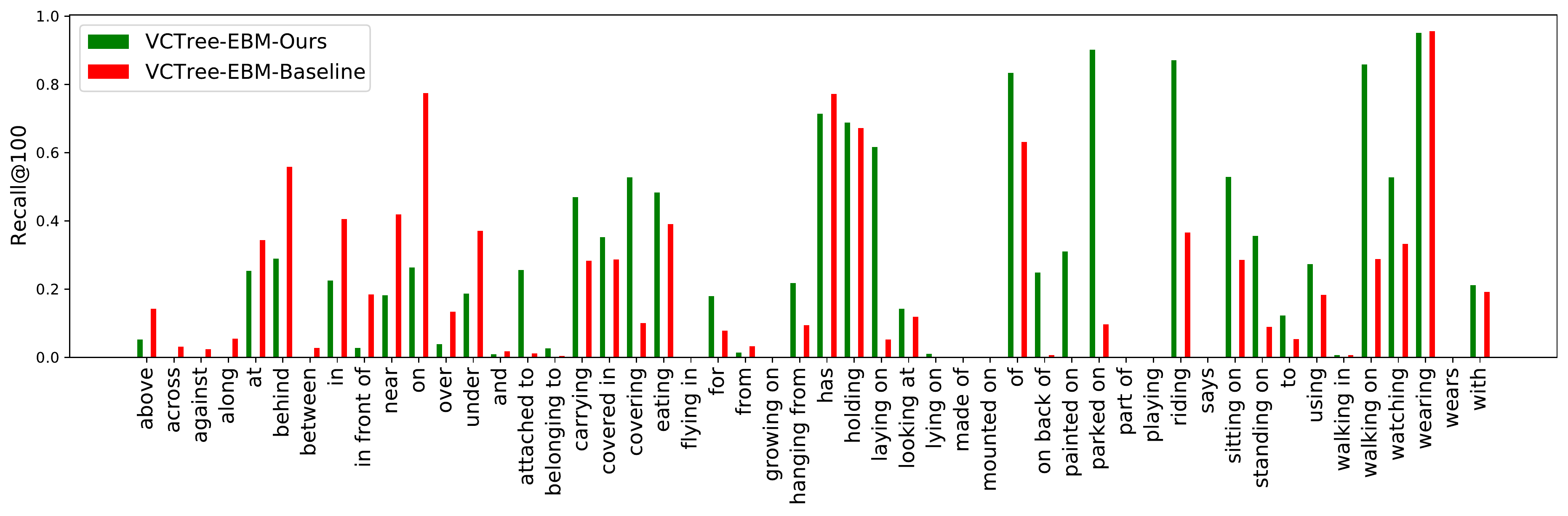}
\end{center}
\caption{Relation-wise Recall using the VCTree-EBM \cite{suhail2021energy} as the backbone SGG model trained with our proposed method (Ours) vs. the method proposed in \cite{suhail2021energy} (Baseline).}
\label{fig:perclasscomparison}
\end{figure*}

\section{Dataset Statistics}
\label{sec.dataset}

For the Visual Genome dataset \cite{krishna2017visual}, Xu \etal \cite{xu2017scene} released a version of the dataset with 50 relations and 150 object categories. These 50 relations are: \emph{above, across, against, along, and, at,  attached to,  behind,  belonging to,  between,  carrying,  covered in, covering, eating,  flying in,  for,  from,  growing on,  hanging from,  has,  holding,  in,  in front of, laying on, looking at,  lying on,  made of,  mounted on,  near,  of,  on,  on back of,  over,  painted on,  parked on, part of,  playing,  riding,  says,  sitting on,  standing on,  to,  under,  using,  walking in,  walking on,  watching,  wearing,  wears,  with.}

In \Cref{table:expset}, we present the explicit set of relations with the number of training instances for each relation in the Visual Genome dataset. Similarly, in \Cref{table:impset1} and \Cref{table:impset2}, we define the implicit set of relations\footnote{We break down the implicit relations into two tables for better visualization.} and their frequency in the Visual Genome dataset.



\section{Additional Results}

\noindent
\textbf{Quantitative Studies. }
In \Cref{fig:intro-fig-full}, we present the class wise recall for all the relation classes in the Visual Genome dataset \cite{krishna2017visual} for training on a subset of relations \ie either learning only on \textit{explicit} relations or only on \textit{implicit} relations. All the models are trained with the MOTIF-TDE-Sum SGG model \cite{zellers2018neural, tang2020unbiased}. The class-wise performances clearly indicates the generalizability of training only on implicit relations as it achieves at-par/similar performances on the explicit relations, whereas the model only trained on explicit relations performs poorly on implicit relations. 

\Cref{fig:perclasscomparison} compares the class-wise performances for the VCTree model trained with only Energy Based Modeling (EBM) \cite{suhail2021energy} and also with our proposed method. The performance gains of our model over the baseline in the implicit relations such as ``carrying'', ``eating'', ``covering'', walking'' \etc shows the importance of mining these informative relations from less informative samples while still maintaining recall on the explicit relations hence, improving generalization.

\noindent
\textbf{Regular Recall Results. }In \Cref{table:sota_rr}, we show the Regular Recall@k results for different SGG backbone architectures when trained with our proposed method compared to the baseline. Although, there is no significant improvement in Regular Recall (when compared to the improvements obtained from mean recall), the at-par performance with the baseline shows that our method maintains the performance on frequent relations while improving significantly on the more informative/infrequent relation classes (as measured by mean recall).

\begin{table*}[!ht]
\renewcommand*{\arraystretch}{1.13}

		\resizebox{0.9\textwidth}{!}{
			\begin{tabular}{c|c |c | c| c| c| c| c| c| c| c}
				\hline
				& & \multicolumn{3}{c|}{Predicate Classification} &\multicolumn{3}{c|}{Scene Graph Classification} & \multicolumn{3}{c}{Scene Graph Detection }\\
				
				 Models & Method & R@20 & R@50 & R@100 & R@20 & R@50 & R@100  & R@20 & R@50 & R@100  \\
				\hline


				
    			\multirow{2}{*}{Motif-TDE-Sum \cite{zellers2018neural, tang2020unbiased}} & Baseline & 33.38 & 45.88 & 51.25 & 20.47 & 26.31 & 28.79 & 11.92 & 16.56 & 20.15 \\ 
				 & Ours  & 33.36 & 43.53 & 47.44 & 24.31 & 29.91 & 31.75 & 14.59 & 17.96 & 19.70 \\ \hline

				\multirow{2}{*}{VCTree \cite{tang2018learning}} & Baseline & 59.82 & 65.93 & 67.57 & 41.49 & 45.16 & 46.10 & 24.90 & 32.02 & 36.30 \\ 
		
				 & Ours &58.66 & 64.69 & 67.05 & 35.49 & 38.71 & 39.51 & 24.63  & 31.52 & 36.42 \\ \hline
				
				\multirow{2}{*}{VCTree-EBM \cite{suhail2021energy}} & Baseline & 57.31 &  63.99 & 65.84 & 40.31 & 44.72 & 45.84 & 24.21 & 31.36 & 35.87   \\ 
			
				 & Ours & 57.42 & 64.37 & 66.43 & 35.42 & 38.79 & 39.66 &23.70 &30.74  &  35.62\\ \hline
				
				\multirow{2}{*}{VCTree-TDE \cite{tang2020unbiased}} & Baseline & 40.12 & 50.83 &54.91 & 26.00 & 33.03 & 35.97 & 13.97 & 19.43 & 23.34 \\

                 & Ours & 36.90 & 47.62  & 52.03  & 25.67 & 32.83 & 35.76 & 15.20 & 19.00 & 20.98 \\ \hline

\end{tabular}}
\caption{Scene Graph Generation performance comparison on Regular Recall@K \cite{tang2020unbiased} under all three experimental settings. We compare the results of our proposed framework (Ours) with the original model (Baseline) using different SGG architectures.}
\label{table:sota_rr}
\end{table*}

\begin{figure*}[!h]
     \centering
     \begin{subfigure}[b]{0.8\textwidth}
         \centering
         \includegraphics[width=\linewidth]{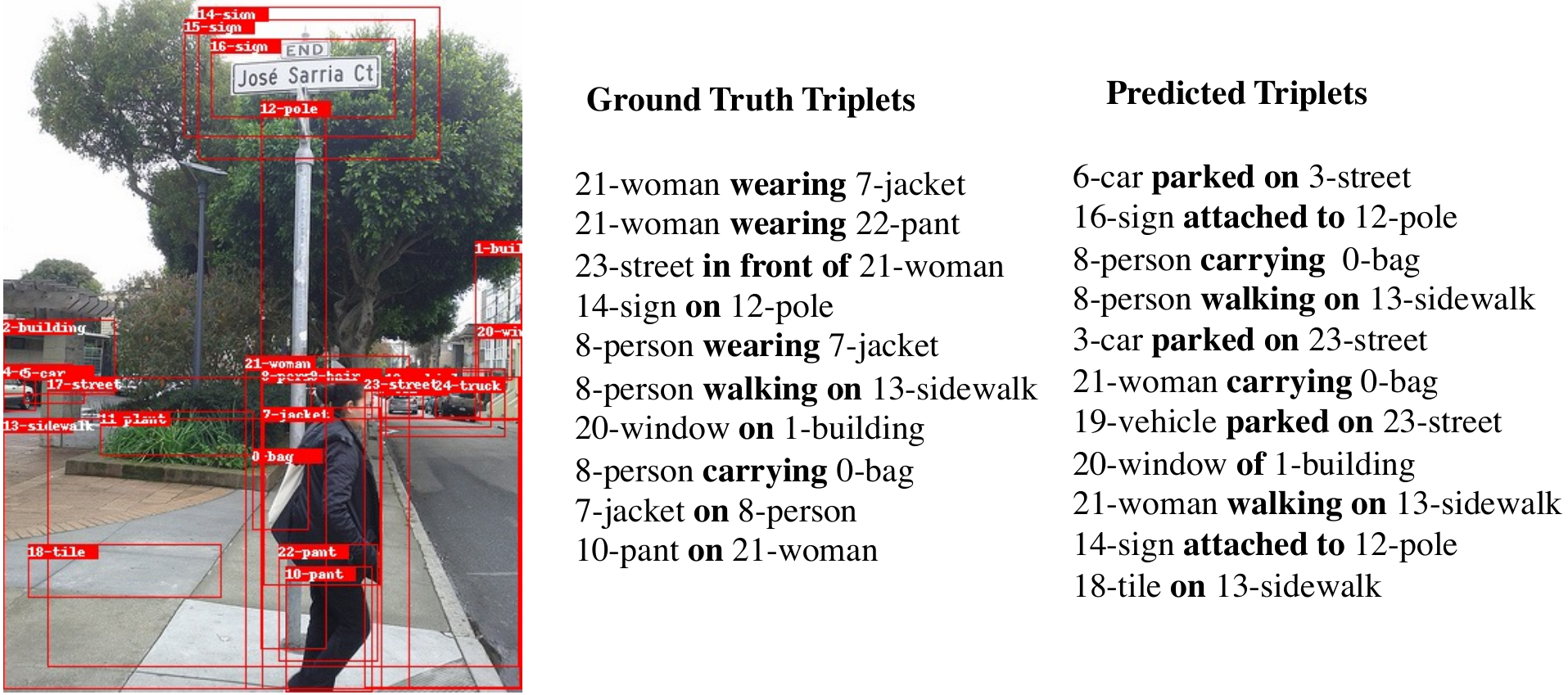}
     \end{subfigure}
     
     \vspace{\floatsep}
     
     \begin{subfigure}[b]{0.8\textwidth}
         \centering
        \includegraphics[width=\linewidth]{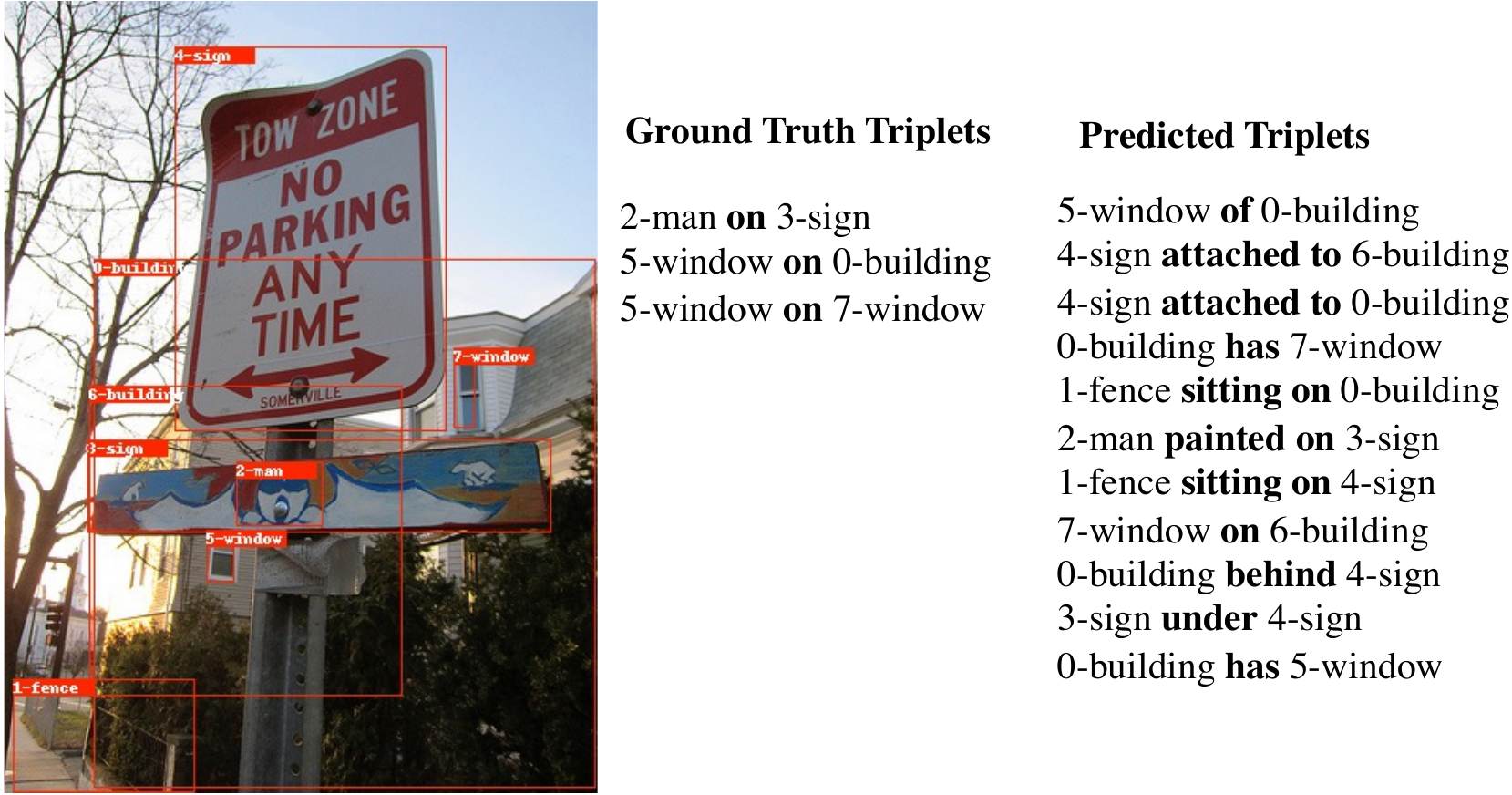}
     \end{subfigure}
        \caption{Additional Qualitative Results with the ground truth triplets and the predicted triplets from the VCTree-EBM model trained with our proposed training framework. The predicted triplets are from the SGCls setting.}
        \label{fig:qual1}
\end{figure*}

\begin{figure*}[!h]
     \centering
     \begin{subfigure}[b]{0.9\textwidth}
         \centering
         \includegraphics[width=\linewidth]{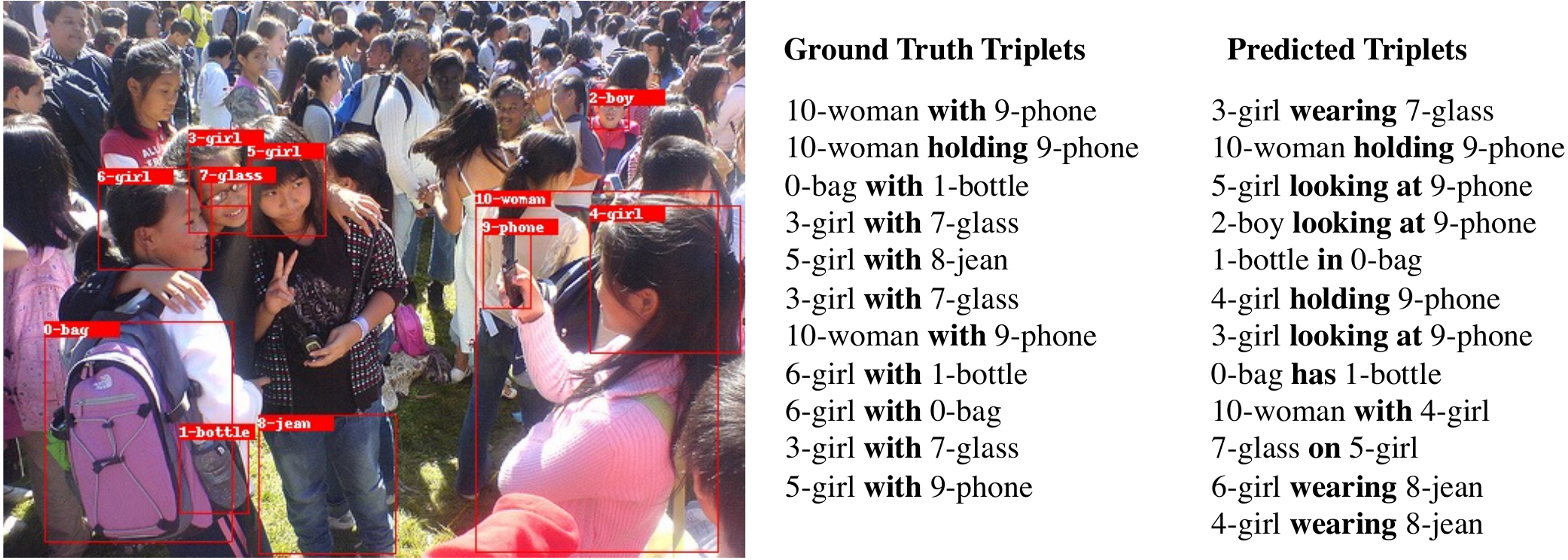}
     \end{subfigure}
     
     \vspace{\floatsep}
     
     \begin{subfigure}[b]{0.9\textwidth}
         \centering
        \includegraphics[width=\linewidth]{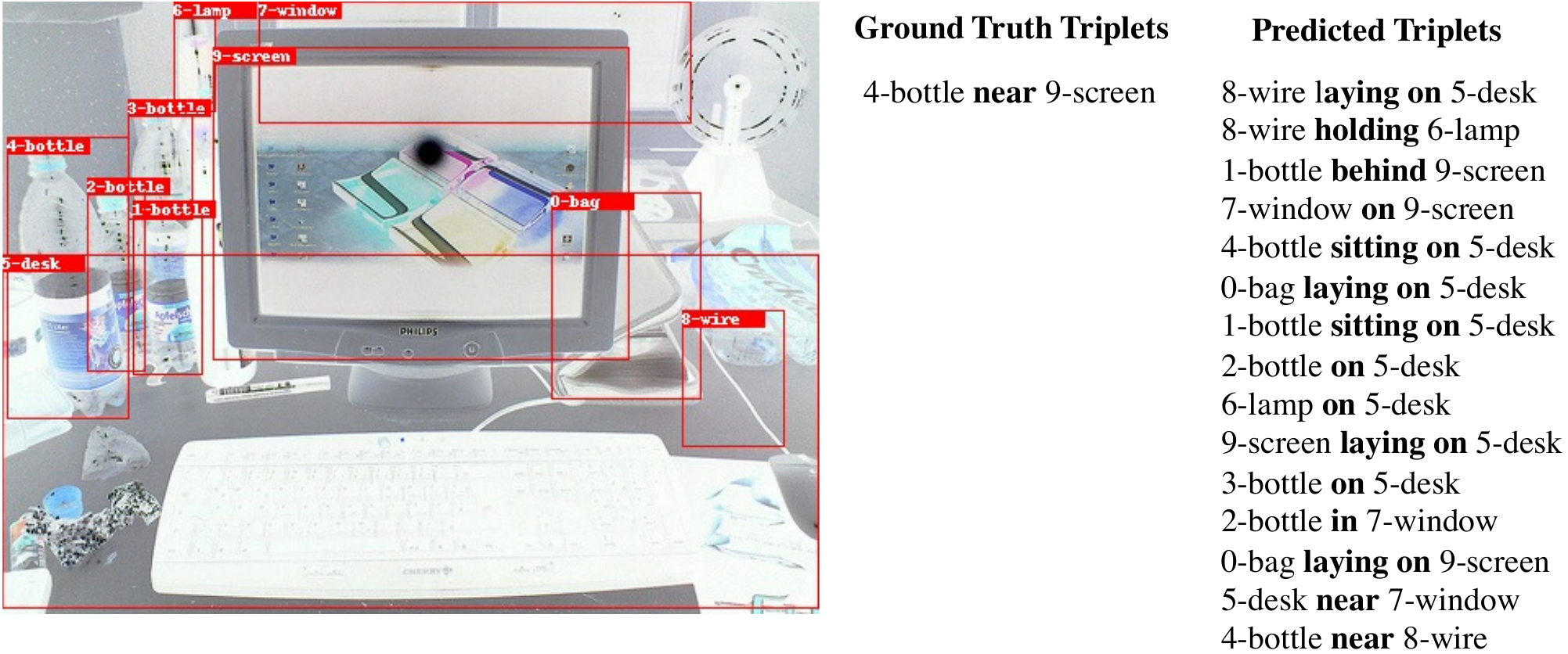}
     \end{subfigure}
        \caption{Additional Qualitative Results.}
        \label{fig:qual2}
\end{figure*}

\noindent
\textbf{Qualitative Studies. }We present additional qualitative visualizations in \Cref{fig:qual1} and \Cref{fig:qual2}. Our proposed method predicts informative relations for both the set of pairs present in the ground truth and new set of object pairs that further helps to define a scene comprehensively. In the quantitative evaluation we only reward object pairs that have corresponding ground truth relations, hence, the relations for the remaining set of object pairs can only be visualized qualitatively.

\clearpage
{\small
\bibliographystyle{ieee_fullname}
\bibliography{refs}
}